\documentclass[conference]{IEEEtran}
\IEEEoverridecommandlockouts

\usepackage{cite}
\usepackage{amsmath,amssymb,amsfonts}
\usepackage{algorithmic}
\usepackage{graphicx}
\usepackage{textcomp}
\usepackage{xcolor}
\usepackage{hyperref}
\def\BibTeX{{\rm B\kern-.05em{\sc i\kern-.025em b}\kern-.08em
    T\kern-.1667em\lower.7ex\hbox{E}\kern-.125emX}}
\begin{document}

\title{DuSCN-FusionNet: An Interpretable Dual-Channel Structural Covariance Fusion Framework for ADHD
Classification Using Structural MRI\\

\thanks{Identify applicable funding agency here. If none, delete this.}
}

\author{
\IEEEauthorblockN{
Qurat Ul Ain\textsuperscript{1,2},
Alptekin Temizel\textsuperscript{2},
Soyiba Jawed\textsuperscript{1}
}

\IEEEauthorblockA{
\textsuperscript{1}Department of Computer and Software Engineering,
College of Electrical and Mechanical Engineering,\\
National University of Sciences and Technology,
Islamabad 44000, Pakistan
}

\IEEEauthorblockA{
\textsuperscript{2}Graduate School of Informatics,
Middle East Technical University, 06800, Ankara, Turkey
}

\IEEEauthorblockA{
qain.cse22ceme@student.nust.edu.pk,
atemizel@metu.edu.tr,
soyiba.jawed@ceme.nust.edu.pk
}
}

\maketitle

\begin{abstract}
Attention Deficit Hyperactivity Disorder (ADHD) is a highly prevalent neurodevelopmental condition; however, its neurobiological diagnosis remains challenging due to the lack of reliable imaging-based biomarkers, particularly anatomical markers. Structural MRI (sMRI) provides a non-invasive modality for investigating brain alterations associated with ADHD; nevertheless, most deep learning approaches function as black-box systems, limiting clinical trust and interpretability. In this work, we propose DuSCN-FusionNet, an interpretable sMRI-based framework for ADHD classification that leverages dual-channel Structural Covariance Networks (SCNs) to capture inter-regional morphological relationships. ROI-wise mean intensity and intra-regional variability descriptors are used to construct intensity-based and heterogeneity-based SCNs, which are processed through an SCN-CNN encoder. In parallel, auxiliary ROI-wise variability features and global statistical descriptors are integrated via late-stage fusion to enhance performance. The model is evaluated using stratified 10-fold cross-validation with a 5-seed ensemble strategy, achieving a mean balanced accuracy of 80.59\% and an AUC of 0.778 on the Peking University site of the ADHD-200 dataset. DuSCN-FusionNet further achieves precision, recall, and F1-scores of 81.66\%, 80.59\%, and 80.27\%, respectively. Moreover, Grad-CAM is adapted to the SCN domain to derive ROI-level importance scores, enabling the identification of structurally relevant brain regions as potential biomarkers.
\end{abstract}


\section{Introduction}
Attention Deficit Hyperactivity Disorder (ADHD) is a common neurodevelopmental disorder that begins in childhood and often persists into adulthood ~\cite{b1}. This is heterogeneous, comprising three subtypes: inattentive (ADHD-I), hyperactive/impulsive (ADHD-HI), and combined (ADHD-C) ~\cite{b3}. Individuals with ADHD frequently experience academic impairment, poorer mental and physical health ~\cite{b4}, and reduced quality of life ~\cite{b5}. In recent decades, neuroimaging has become an essential tool for investigating brain alterations and supporting clinical diagnosis ~\cite{b6}. In particular, structural MRI (sMRI) provides valuable insights into neuroanatomical abnormalities associated with neurological and psychiatric disorders, including ADHD ~\cite{b7}.
In recent years , researchers start utilizing traditional machine learning (ML) approaches for ADHD diagnosis. For instance, Zhang-James et al.~\cite{b8} applied ML models to sMRI data from both youth and adult cohorts, identifying shared structural anomalies across age groups. Lohani et al.~\cite{b9} combined sMRI with personal characteristic features from the ADHD-200 dataset and employed mRMR and exhaustive feature selection using SVM-based classifiers. Similarly, Sachnev et al.~\cite{b10} extracted regional sMRI features, optimized feature subsets via a genetic algorithm, and utilized extreme learning machines for classification. More recently, DL approaches have gained substantial attention due to their ability to automatically learn hierarchical and discriminative representations from neuroimaging data~\cite{b11}. Peng et al.~\cite{b12} proposed a dual 3D-CNN framework integrating sMRI and fMRI, while Wang et al.~\cite{b13} introduced a multi-scale attention-based 3D-CNN for mental disorder diagnosis including ADHD. Furthermore, Kuttala et al.~\cite{b14} developed a dense attentive GAN-based one class model trained on healthy sMRI scans to detect deviations associated with ADHD.
\par
Although recent studies have applied ML and DL to sMRI for ADHD diagnosis, this modality still remains relatively under-explored ~\cite{b15}. Moreover, despite their strong performance, DL models are often limited in clinical practice by their black-box nature, which obscures the anatomical basis of predictions. Interpretability is therefore essential for reliable medical decision-making. While explainable AI (XAI) methods such as CAM and Grad-CAM ~\cite{b16} have been widely used in other imaging applications, their adoption for ADHD classification from sMRI is still limited ~\cite{b15}. This underscores the need for interpretable sMRI-based frameworks that not only accurately distinguish ADHD from healthy controls (HCs), but also localize disorder-relevant brain regions as potential biomarkers. Therefore, we propose an interpretable sMRI-driven ADHD classification framework that enables both accurate diagnosis and anatomical localization. The main contributions of this work are as follows:
\begin{itemize}
    \item We develop an sMRI-only deep learning pipeline for ADHD identification based on dual-channel Structural Covariance Network (SCN) representations.

    \item We encode intensity-based and heterogeneity-based SCNs via an SCN-CNN branch, while auxiliary ROI variability and global statistical features are integrated through late-stage fusion for improved ADHD--HC discrimination.

    \item We adapt Grad-CAM to the SCN-CNN module to compute ROI-level importance scores, enabling localization of structurally relevant brain regions as candidate neuroanatomical biomarkers associated with ADHD.
\end{itemize}

\section{Proposed Methodology}
\begin{figure*}[t]
    \centering
    \includegraphics[width=0.85\textwidth]{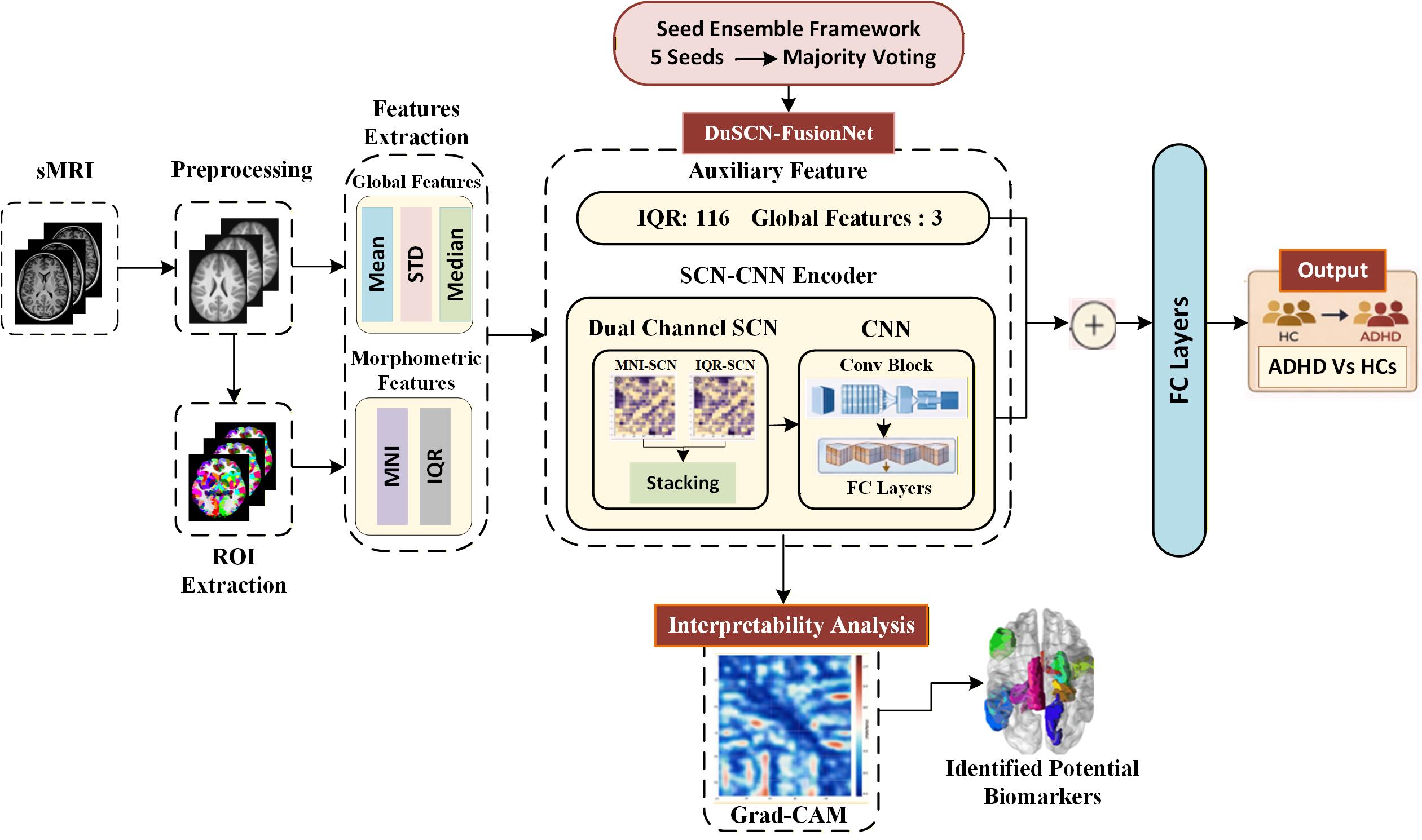}
    \caption{Overview of the proposed DuSCN-FusionNet framework for ADHD classification.}
    \label{fig:overview_framework}
\end{figure*}

In this study, we propose DuSCN-FusionNet (Dual-channel SCN-CNN Fusion Network), a deep learning framework for ADHD classification using sMRI. After data preprocessing, the network consists of three modules: Feature Extraction, Structural Covariance Networks (SCNs) Modeling, and Classification. Post-hoc interpretability is applied to identify the brain regions and features most influential in the predictions. The overall architecture is shown in Fig.~\ref{fig:overview_framework}.

\subsection{Data Preprocessing}\label{AA}
In this study, T1 weighted MRI scans from the ADHD-200 dataset, preprocessed using the Athena pipeline ~\cite{b17} are utilized. Specifically, skull stripped and spatially normalized images are employed. To mitigate inter subject intensity variability and scanner related bias, a robust intensity normalization procedure is applied. For each subject, voxel intensities within the brain mask are normalized using the median and median absolute deviation (MAD). The resulting normalized values are subsequently clipped to the range [-3,3] and linearly rescaled to [0,1].  

\subsection{Feature Extraction Module}\label{AA}

Region of Interest (ROI)-based analysis is performed, which divides the brain into distinct regions and extracts valuable information from each ROI. The Automated Anatomical Labeling (AAL) atlas ~\cite{b18}, comprising 116 ROIs, is used to parcel the normalized structural images. The atlas is resampled to match the image resolution using nearest-neighbor interpolation to preserve discrete labels. \par
For each ROI $r$, two complementary morphometric features are extracted from the normalized image: (1) mean normalized intensity (MNI), and (2) inter-quartile range (IQR), as defined in Eq.~\ref{eq:roi_mean} and Eq.~\ref{eq:iqr}. The ROI-wise mean intensity is computed as

\begin{equation}
\mu_r = \frac{1}{|V_r|} \sum_{v \in V_r} x_v
\label{eq:roi_mean}
\end{equation}

where $r$ indexes the ROI, $V_r$ denotes the set of voxels belonging to ROI $r$, $|V_r|$ is the number of voxels in that region, $v$ represents an individual voxel index, and $x_v$ is the normalized intensity value at voxel $v$.

\begin{equation}
\mathrm{IQR}_r = Q_{75}(V_r) - Q_{25}(V_r)
\label{eq:iqr}
\end{equation}
The interquartile range in (\ref{eq:iqr}) captures intra-regional structural heterogeneity. Additionally, three global image statistics (mean, standard deviation, and median) are computed as auxiliary features.

\subsection{Structural Covariance Networks Modeling Module}\label{AA}
SCNs represent inter-regional relationships by modeling similarities in brain morphology across subjects using sMRI and have been widely used to characterize large-scale brain organization in both healthy populations and neurological disorders ~\cite{b19}. In this work, SCNs are constructed from ROI-wise morphometric features extracted after AAL parcellation.  

1) \textbf{Group-Level Structural Covariance:}  Let $\mathbf{F} \in \mathbb{R}^{N \times R}$ denote the feature matrix of $N$ training subjects, where each row corresponds to a subject and each column corresponds to one of $R = 116$ AAL regions.The group-level structural covariance matrix is computed in Eq.~\ref{a:b}.
\begin{equation}
\mathbf{C} = \mathrm{corr}(\mathbf{F})
\label{a:b}
\end{equation} 
encoding population-level inter-regional structural associations.  

2) \textbf{Subject-Specific Structural Covariance:} To preserve individual variability, the ROI feature vector of subject $s$ is $f(s) \in \mathbb{R}^{R}$. The normalized outer-product representation is given in Eq.~\ref{c:d}.
\begin{equation}
\mathbf{C}_{\text{ind}}(s)
=
\left(
\frac{f(s)}{\lVert f(s) \rVert_2}
\right)
\left(
\frac{f(s)}{\lVert f(s) \rVert_2}
\right)^{\top}
\label{c:d}
\end{equation}
The final subject-specific SCN blends group-level and individual-level information presented in  Eq.~\ref{e:f}.
\begin{equation}
\mathbf{C}(s) = \alpha \mathbf{C} + (1 - \alpha)\mathbf{C}_{\text{ind}}(s)
\label{e:f}
\end{equation}

3) \textbf{Dual-Channel SCN Representation:} Two SCNs are constructed per subject: (i) an intensity-based SCN from ROI-wise mean normalized intensities, and (ii) a heterogeneity-based SCN from ROI-wise IQR features. Both channels use the same group and subject-specific blending strategy. The resulting matrices are stacked to form a dual-channel SCN tensor, used as input to the CNN: $\mathbf{X}(s) \in \mathbb{R}^{2 \times 116 \times 116}$.

\subsection{Classification Module}\label{AA}
We employ a convolutional neural network (CNN) to classify ADHD and HC subjects using each subject’s dual-channel SCN as input. The network has three convolutional layers with 64, 128, and 256 filters, each followed by batch normalization and ReLU activation. Max pooling is applied after the first two layers, and adaptive average pooling follows the final convolution to produce a compact feature vector, which is then passed through two fully connected (FC) layers.

A vector of 119 auxiliary structural features, including ROI-wise IQR and three global statistics, is processed through two FC layers. The CNN-derived SCN vector and auxiliary feature vector are concatenated and passed through two additional FC layers, culminating in a final classification layer as defined in Eq.~\ref{eq:prediction}.
\begin{equation}
\hat{y}(s)
=
\mathrm{softmax}
\left(
\mathbf{W}
\left[
\mathbf{z}_{\mathrm{SCN}}(s) \,\|\, \mathbf{z}_{\mathrm{aux}}(s)
\right]
+
\mathbf{b}
\right)
\label{eq:prediction}
\end{equation}

where $\mathbf{z}_{\mathrm{SCN}}(s)$ is the CNN-derived SCN feature vector, $\mathbf{z}_{\mathrm{aux}}(s)$ denotes auxiliary structural features, $\lVert  $ indicates concatenation, and $\mathbf{W}$ and $\mathbf{b}$ are learnable parameters. Post-hoc interpretability is then applied to identify brain regions contributing to ADHD classification.

\subsection{Model Interpretability (Post-hoc Analysis)} 
To identify structurally relevant brain regions contributing to ADHD classification, post-hoc interpretability is performed on the SCN-CNN branch using Gradient-weighted Class Activation Mapping (Grad-CAM) ~\cite{b16}. Gradients of the ADHD class score are back-propagated to the final convolutional layer to compute activation maps. Since SCNs encode inter-regional relationships rather than voxel-wise spatial patterns, Grad-CAM responses are aggregated across rows and columns of the SCN adjacency matrix to derive ROI-level importance scores. Scores are averaged across subjects and ROIs exceeding the 90th percentile are retained as candidate structural biomarkers; this threshold is empirically chosen to emphasize highly
discriminative responses while suppressing diffuse activations. In addition, ROI-wise statistical testing is conducted to independently assess group-level anatomical differences between ADHD and HCs.

\section{Experimental Evaluation}
\subsection{Data Acquisition}\label{AA}
Structural MRI scans have been acquired from the Peking University (PU) site of the publicly available ADHD-200 
\footnote{\url{https://neurobureau.projects.nitrc.org/ADHD200/Introduction.html}} consortium
, comprising 194 subjects (116 healthy controls and 78 ADHD). 
\subsection{Classification Performance}\label{AA}
The classification performance of DuSCN-FusionNet is evaluated using stratified 10-fold cross-validation in a binary setting. The model is trained with the Adam optimizer (initial learning rate $1\times10^{-4}$) for up to 50 epochs, with early stopping triggered after 15 epochs without validation improvement. A batch size of 4 is used, and dropout regularization (0.2, 0.3) is incorporated within both the SCN-CNN encoder and the auxiliary feature branch prior to feature fusion to mitigate overfitting. The structural covariance blending parameter is fixed to $\alpha = 0.55$, as determined empirically during model development.
\par
\begin{figure}[t]
    \centering
    \includegraphics[width=0.65\linewidth]{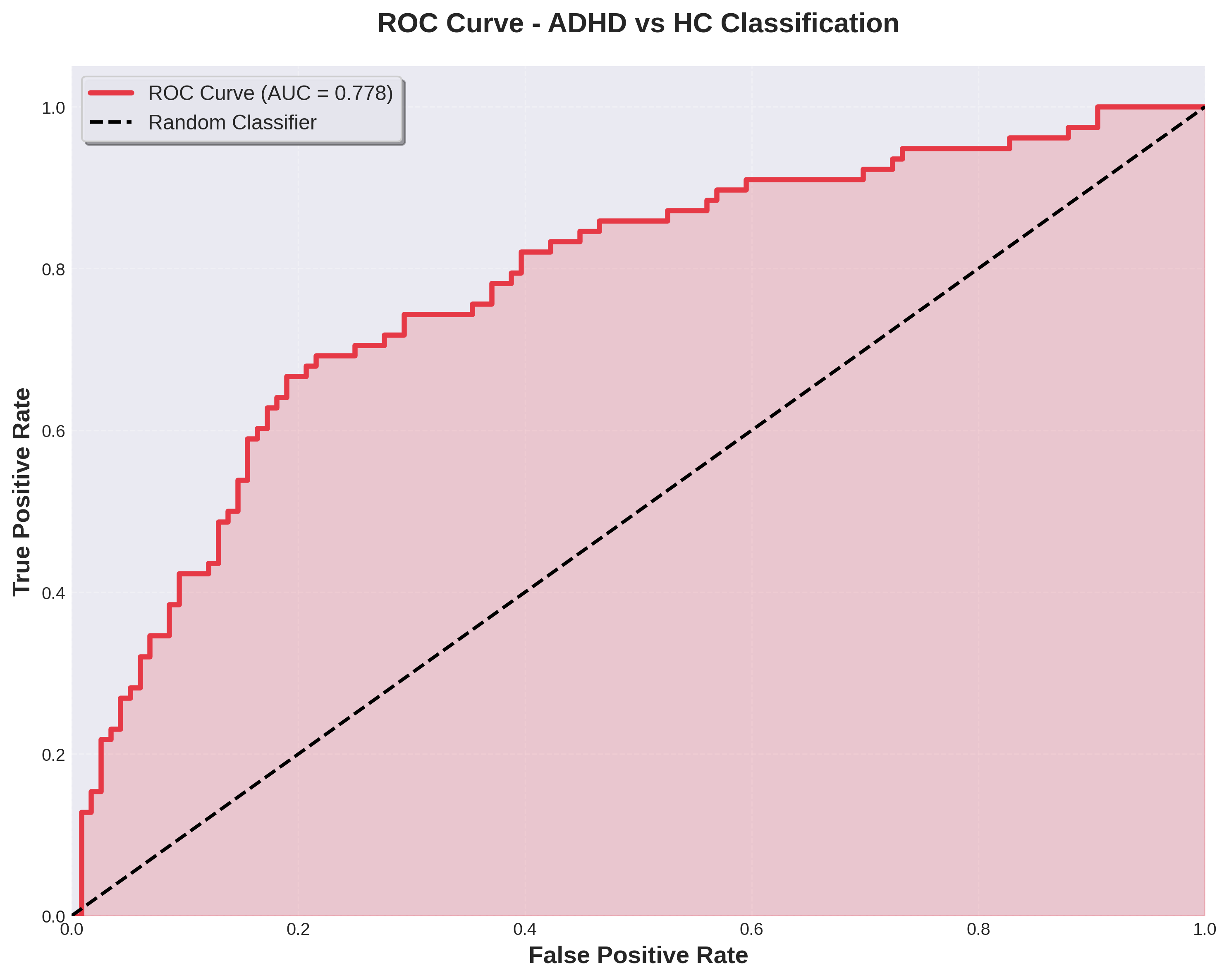}
    \caption{ROC curve for DuSCN-FusionNet in ADHD vs.\ HC classification (AUC = 0.778), outperforming random chance.}
    \label{fig:area}
\end{figure}
\begin{figure}[t]
    \centering
    \includegraphics[width=0.65\linewidth]{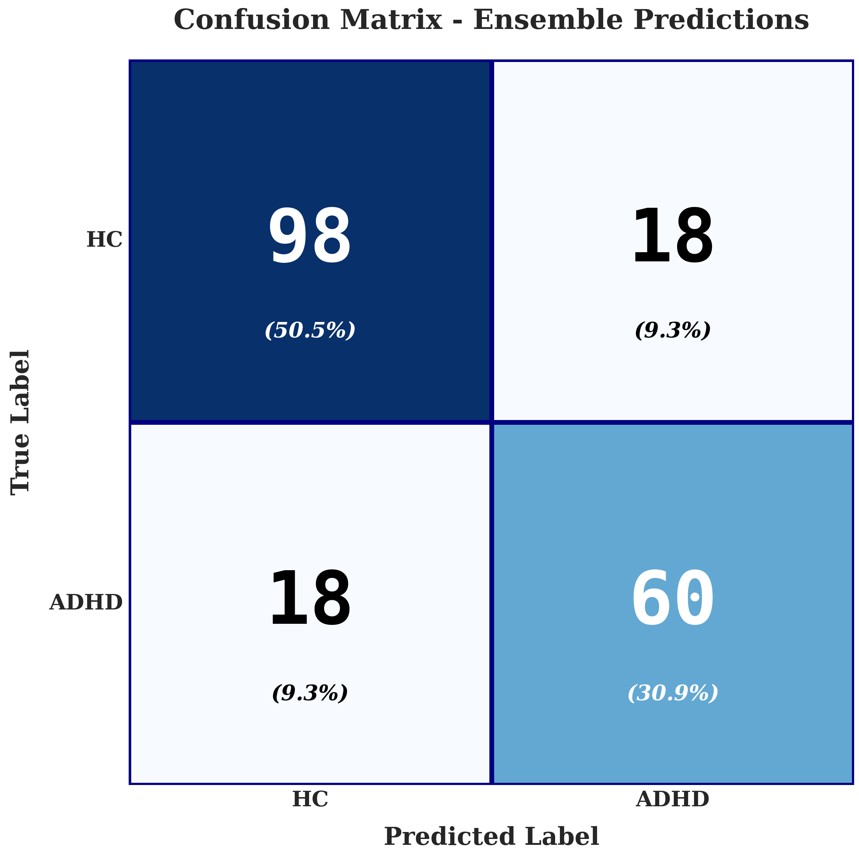}
    \caption{Confusion matrix of DuSCN-FusionNet predictions for ADHD and HCs, summarizing classification performance across subjects.}
    \label{fig:cm}
\end{figure}

To improve robustness and reproducibility, a seed-based ensemble strategy is adopted, where the same architecture is trained with different random seeds ~\cite{b20}. Specifically, a 5-seed ensemble is employed within each fold, and final predictions are obtained via majority voting. Performance is measured using balanced accuracy (ACC) due to class imbalance, along with precision (PRE), recall (REC), F1-score (F1), and AUC (Fig.~\ref{fig:area}). Across folds, DuSCN-FusionNet achieved a mean balanced accuracy of $80.59\%\pm9.39$ and an AUC of 0.778. The confusion matrix (Fig.~\ref{fig:cm}) yielded precision, recall, and F1-score of $81.66\%\pm8.87$, $80.59\%\pm9.39$, and $80.27\%\pm9.28$, respectively, demonstrating reliable discrimination between ADHD and HCs.

\subsection{Interpretability Results }\label{AA}
\begin{figure}[t]
    \centering
    \includegraphics[width=0.65\linewidth]{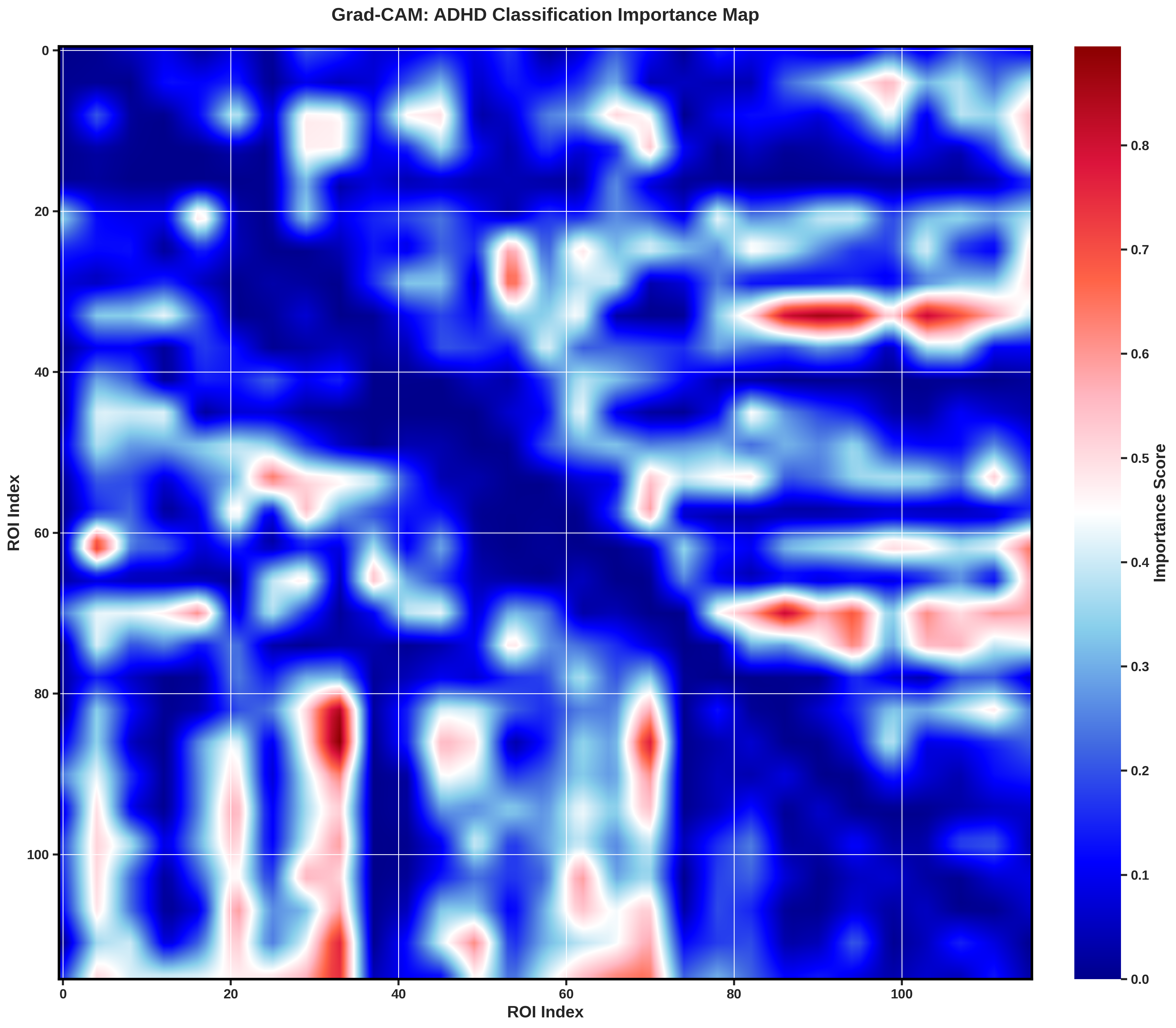}
    \caption{ROI-level interpretability analysis using Grad-CAM on the SCN-CNN branch.}
    \label{fig:gradcam_roi_map}
\end{figure}

\begin{figure}[t]
\centering
\includegraphics[width=0.90\linewidth]{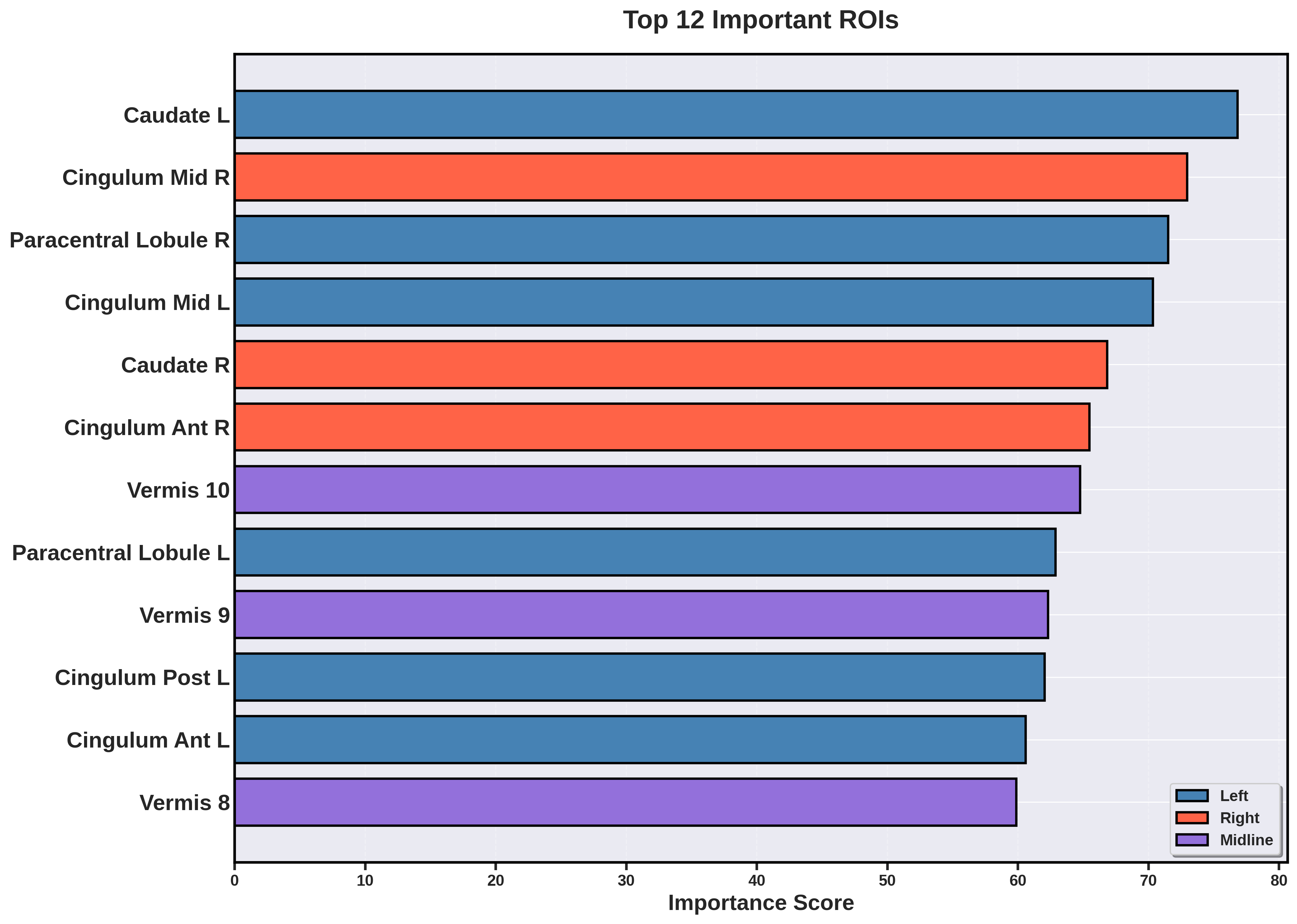}
\caption{Top 12 most salient brain regions identified by Grad-CAM. 
ROIs above the 90th percentile threshold are retained as candidate structural biomarkers for ADHD classification.}
\label{fig:top12_rois}
\end{figure}

Grad-CAM analysis (Fig.~\ref{fig:gradcam_roi_map}) on the best-performing cross-validation fold revealed that the SCN-CNN model focuses on a limited set of highly discriminative structural network interactions. Using a 90th-percentile threshold, 12 suprathreshold ROIs have been identified (Fig.~\ref{fig:top12_rois}), with strongest contributions from the bilateral caudate, cingulum (anterior/middle/posterior), paracentral lobules, and cerebellar vermis (Vermis 9–10). These regions align with established ADHD neurobiology: the fronto-striatal circuit, including the caudate, is critical for executive control and attention regulation ~\cite{b21}; cingulate abnormalities relate to impaired cognitive control and emotion processing ~\cite{b22}; and the cerebellar vermis contributes to attentional and affective modulation deficits ~\cite{b23}.
To support these model-derived findings, complementary ROI-wise statistical testing further confirmed significant group differences across multiple AAL regions after Bonferroni/FDR correction, with 61 ROIs showing large effects (|d|$>$0.8). Network-level comparisons also indicated coordinated structural alterations (Kruskal–Wallis H = 18.01, p = 0.021), suggesting widespread ADHD-related disruptions consistent with the Grad-CAM importance patterns.

\begin{table}[h!]
\centering
\caption{Comparison of ADHD classification performance with previous studies and the proposed DuSCN-FusionNet pipeline.}
\label{tab:comparison}
\setlength{\tabcolsep}{4pt}
\renewcommand{\arraystretch}{1.1}
\footnotesize
\begin{tabular}{|p{1.4cm}|p{1.4cm}|p{1.7cm}|p{1.3cm}|p{0.8cm}|p{0.9cm}|p{1.4cm}|}
\hline
\textbf{Method}  & \textbf{Sample (ADHD/HC)} & \textbf{Features} & \textbf{Classifier(s)} & \textbf{ACC (\%)} & \textbf{Bio-markers} \\
\hline
Wang et al. [12]  & 146 / 441 & Raw sMRI & 3D MVA-CNN & 78.8 & No \\
\hline
Kuttala et al. [13]  & 77 / 94 & Multilayer Perceptron & Dense Attentive GAN & 85.38 & No \\
\hline
Zou et al. [23]  & 197 / 362 & GM, WM, CSF & 3D CNN & 65.86 & No \\
\hline
\textbf{DuSCN-FusionNet-(ours)}  & 78 / 116 & SCNs + IQR + Global Features & CNN-MLP & 80.59 & Yes \\
\hline
\end{tabular}
\end{table}

\begin{table}[h!]
\centering
\caption{Ablation study of the DuSCN-FusionNet pipeline showing classification performance.}
\label{tab:ablation}
\footnotesize
\begin{tabular}{|p{2cm}|p{1.2cm}|p{1.2cm}|p{1.2cm}|p{1.2cm}|}
\hline
\textbf{\shortstack{Pipeline \\ Variant}}& \textbf{\shortstack {ACC (\%)}} & \textbf{\shortstack {PRE (\%)}} & \textbf{REC (\%)} & \textbf{F1 (\%)} \\
\hline
Without Auxiliary features  & 62.01 $\pm$ 13.75
 & 63.57 $\pm$ 14.59
 & 62.01 $\pm$ 13.75 & 60.03 $\pm$ 13.30 \\
\hline
Without Seed Ensemble & 72.14 $\pm$ 6.17 & 74.04 $\pm$ 6.17
 & 72.14 $\pm$ 6.17 & 71.50 $\pm$ 6.52 \\
\hline
DuSCN-FusionNet & 80.59 $\pm$ 9.39 & 81.66 $\pm$ 8.87 & 80.59 $\pm$ 9.39 & 80.27 $\pm$ 9.28 \\
\hline
\end{tabular}
\end{table}

\subsection{Performance Comparison}\label{AA}
The performance of the proposed DuSCN-FusionNet is compared with several state-of-the-art (SOTA) methods that employ sMRI data from the ADHD-200 consortium (Table~\ref{tab:comparison}).Wang et al. ~\cite{b13} utilized raw structural MRI data with a 3D MVA-CNN on a large and imbalanced cohort (146 ADHD / 441 HC), achieving 78.8\% accuracy without model interpretation.
Kuttala et al. ~\cite{b14} achieved higher accuracy (85.38\%) on 77/94 subjects using a dense attentive GAN; however, the approach lacks interpretability and biomarker-level insights. Zou et al. ~\cite{b24} employed gray matter, white matter, and CSF representations with a 3D CNN on a larger cohort (197 / 362), yielding 65.86\% accuracy. Using SCN representations derived from ROI-level intensity and variability descriptors combined with IQR and global statistical features, DuSCN-FusionNet achieves competitive performance (80.59\%) on a moderate-sized subset (78 / 116) while additionally providing ROI-level potential biomarker interpretability via Grad-CAM analysis.Direct numerical comparison should be treated cautiously because the studies use different sample sizes and potentially different ADHD-200 sites, which may affect the comparison. Nevertheless, the proposed framework demonstrates competitive discrimination while providing interpretability not offered by the compared methods.

\subsection{Ablation Study}\label{AA}
An ablation study has been conducted to evaluate the DuSCN-FusionNet pipeline (Table~\ref{tab:ablation}). The full model, incorporating the structural covariance networks (SCNs), auxiliary structural features, and seed ensemble, achieved the highest performance across all metrics.

\section{Conclusion and Future Work}

This work introduces DuSCN-FusionNet, an interpretable structural MRI–driven framework that models inter-regional morphological relationships through dual-channel Structural Covariance Networks and integrates complementary information via late fusion of ROI variability features and global statistical descriptors. Beyond achieving competitive classification performance on the ADHD-200 Peking University cohort, the proposed design demonstrates that explicitly modeling structural covariance alongside intra-regional heterogeneity provides discriminative representations while preserving anatomical interpretability. The adaptation of Grad-CAM to the SCN domain further enables identification of neuroanatomically meaningful regions, linking model predictions with established ADHD-related circuitry and reinforcing the potential of explainable AI for clinically relevant neuroimaging analysis.

Future work will focus on validating generalizability across multi-site ADHD-200 cohorts, incorporating richer morphometric descriptors, and exploring advanced explainability strategies to strengthen biomarker reliability and clinical translation.


\begin{thebibliography}{00}
\bibitem{b1} M. L. Bitsko, R. H. Ghandour, R. M. Holbrook, J. R. Kogan, M. D. Blumberg, and S. J. Blumberg, 
``Prevalence of parent-reported ADHD diagnosis and associated treatment among US children and adolescents,'' 
\emph{J. Clin. Child Adolesc. Psychol.}, vol. 47, no. 2, pp. 199--212, 2018.



\bibitem{b3} 
American Psychiatric Association, 
``Diagnostic and statistical manual of mental disorders,'' 
5th ed., Am. Psychiatr. Assoc., 2013, pp. 591--64.


\bibitem{b4} R. G. Klein, S. Mannuzza, M. A. Ramos Olazagasti, E. Roizen, J. A. Hutchison, E. C. Lashua, and F. X. Castellanos, 
``Clinical and functional outcome of childhood attention-deficit/hyperactivity disorder 33 years later,'' 
\emph{Arch. Gen. Psychiatry}, vol. 69, no. 12, pp. 1295--1303, 2012.

\bibitem{b5} R. Agarwal, M. Goldenberg, R. Perry, and W. W. IsHak, 
``The quality of life of adults with attention deficit hyperactivity disorder: A systematic review,'' 
\emph{Innov. Clin. Neurosci.}, vol. 9, no. 5--6, p. 10, 2012.



\bibitem{b6} J. Zhang, K. Chen, D. Wang, F. Gao, Y. Zheng, and M. Yang, 
``Advances of neuroimaging and data analysis,'' 
\emph{Front. Neurol.}, vol. 11, p. 257, 2020.



\bibitem{b7} T. Frodl and N. Skokauskas, 
``Meta-analysis of structural MRI studies in children and adults with ADHD indicates treatment effects,'' 
\emph{Acta Psychiatr. Scand.}, vol. 125, no. 2, pp. 114--126, 2012.


\bibitem{b8} Y. Zhang-James, E. C. Helminen, J. Liu, B. Franke, M. Hoogman, and S. V. Faraone, 
``Evidence for similar structural brain anomalies in youth and adult attention-deficit/hyperactivity disorder: A machine learning analysis,'' 
\emph{Transl. Psychiatry}, vol. 11, no. 1, p. 82, 2021.
\bibitem{b9} D. C. Lohani and B. Rana, 
``ADHD diagnosis using structural brain MRI and personal characteristic data with machine learning framework,'' 
\emph{Psychiatry Res. Neuroimaging}, vol. 334, p. 111689, 2023.
\bibitem{b10} V. Sachnev and B. S. Mahanand, 
``Efficient ADHD diagnosis system based on structural MRI and specially tailored machine learning technique,'' 
\emph{J. Digit. Contents Soc.}, vol. 25, no. 2, pp. 475--483, 2024.
\bibitem{b11} B. Zhao, C. Cheng, Z. Peng, X. Dong, and G. Meng, 
``Detecting early damages in structures with nonlinear output frequency response functions and the CNN-LSTM model,'' 
\emph{IEEE Trans. Instrum. Meas.}, vol. 69, no. 12, pp. 9557--9567, 2020.

\bibitem{b12} J. Peng, M. Debnath, and A. K. Biswas, 
``Efficacy of novel summation-based synergetic artificial neural network in ADHD diagnosis,'' 
\emph{Mach. Learn. Appl.}, vol. 6, p. 100120, 2021.

\bibitem{b13} Z. Wang, Y. Zhu, H. Shi, Y. Zhang, and C. Yan, 
``A 3D multiscale view convolutional neural network with attention for mental disease diagnosis on MRI images,'' 
\emph{Math. Biosci. Eng.}, vol. 18, pp. 6978--6994, 2021.

\bibitem{b14} D. Kuttala, D. Mahapatra, R. Subramanian, and V. R. M. Oruganti, 
``Dense attentive GAN-based one-class model for detection of autism and ADHD,'' 
\emph{J. King Saud Univ.-Comput. Inf. Sci.}, vol. 34, no. 10, pp. 10444--10458, 2022.

\bibitem{b15} Q. U. Ain, S. Jawed, A. R. Subhani, M. U. Akram, and W. H. Butt, 
``Examining AI-powered ADHD diagnosis: Current trends, key challenges, and future directions in the field,'' 
\emph{IEEE Access}, 2025.
\bibitem{b16} D. Tang, J. Chen, L. Ren, X. Wang, D. Li, and H. Zhang, 
``Reviewing CAM-based deep explainable methods in healthcare,'' 
\emph{Appl. Sci.}, 2023.
\bibitem{b17}
P. Bellec, C. Chu, F. Chouinard‑Decorte, Y. Benhajali, D. S. Margulies, and R. C. Craddock, 
``The Neuro Bureau ADHD‑200 Preprocessed repository,'' 
\emph{NeuroImage}, vol. 144, pp. 275–286, 2017, doi:10.1016/j.neuroimage.2016.06.034.
\bibitem{b18}
N. Tzourio-Mazoyer, B. Landeau, D. Papathanassiou, F. Crivello, O. Etard, N. Delcroix, B. Mazoyer, and M. Joliot, 
``Automated anatomical labeling of activations in SPM using a macroscopic anatomical parcellation of the MNI MRI single-subject brain,''
\emph{NeuroImage}, vol. 15, no. 1, pp. 273--289, 2002, doi:10.1006/nimg.2001.0978.
\bibitem{b19}
N. Mongay‑Ochoa, G. Gonzalez‑Escamilla, V. Fleischer, D. Pareto, À. Rovira, J. Sastre‑Garriga, and S. Groppa, 
``Structural covariance analysis for neurodegenerative and neuroinflammatory brain disorders,''
\emph{Brain}, vol. 148, no. 9, pp. 3072--3084, May 2025, doi:10.1093/brain/awaf151. 
.
\bibitem{b20}
D.-A. Sterpu, D. Măriuța, G. Cican, C.-M. Larco, and L.-T. Grigorie, 
``Machine learning prediction of airfoil aerodynamic performance using neural network ensembles,''
\emph{Applied Sciences}, vol. 15, no. 14, p. 7720, 2025.


\bibitem{b21} S. Cortese, C. Kelly, C. Chabernaud, E. Proal, A. Di Martino, M. P. Milham, and F. X. Castellanos, 
``Toward systems neuroscience of ADHD: A meta-analysis of 55 fMRI studies,'' 
\emph{Am. J. Psychiatry}, vol. 169, no. 10, pp. 1038--1055, 2012.
\bibitem{b22} G. Bush, K. Luu, and M. I. Posner, 
``Cingulate cortex and ADHD: Cognitive and emotional dysfunction,'' 
\emph{CNS Spectrums}, vol. 10, no. 9, pp. 724--731, 2005.
\bibitem{b23} C. J. Stoodley, 
``The cerebellum and cognition: Evidence from functional imaging studies,'' 
\emph{The Cerebellum}, vol. 11, no. 2, pp. 352--365, 2012.

\bibitem{b24} L. Zou, J. Zheng, C. Miao, M. J. McKeown, and Z. J. Wang, 
``3D CNN based automatic diagnosis of attention deficit hyperactivity disorder using functional and structural MRI,'' 
\emph{IEEE Access}, vol. 5, pp. 23626--23636, 2017.


\end{thebibliography}
\end{document}